\documentclass[letterpaper, 10 pt, conference]{ieeeconf}  

\IEEEoverridecommandlockouts                              

\usepackage{graphics} 
\usepackage{epsfig} 
\usepackage{amsmath} 
\usepackage{amssymb}  
\usepackage{breqn} 
\usepackage{booktabs}
\usepackage{algorithm}
\usepackage{algorithmic}
\usepackage[hyperfootnotes=false]{hyperref}
\AtBeginDocument{%
	\let\oldref\ref%
	\def\ref{\oldref*}}

\graphicspath{{fig/}}

\title{\LARGE \bf
SGD for robot motion? The effectiveness of stochastic optimization on a new benchmark for biped locomotion tasks
}

\author{Martim Brand\~{a}o, Kenji Hashimoto and Atsuo Takanishi
\thanks{*This work was supported by ImPACT TRC Program of Council for Science, Technology and Innovation (Cabinet Office, Government of Japan).}
\thanks{M. Brand\~{a}o is with the Research Institute for Science and Engineering, Waseda University: \#41-304, 17 Kikui-cho, Shinjuku-ku, Tokyo 162-0044, JAPAN.}
\thanks{K. Hashimoto is with the Waseda Institute for Advanced Study, and is a researcher at the Humanoid Robotics Institute (HRI).}
\thanks{A. Takanishi is with the Department of Modern Mechanical Engineering, Waseda University; and the director of the Humanoid Robotics Institute (HRI), Waseda University.}%
}

\begin{document}

\maketitle
\thispagestyle{empty}
\pagestyle{empty}

\begin{abstract}

Trajectory optimization and posture generation are hard problems in robot locomotion, which can be non-convex and have multiple local optima. 
Progress on these problems is further hindered by a lack of open benchmarks, since comparisons of different solutions are difficult to make.

In this paper we introduce a new benchmark for trajectory optimization and posture generation of legged robots, using a pre-defined scenario, robot and constraints, as well as evaluation criteria.
We evaluate state-of-the-art trajectory optimization algorithms based on sequential quadratic programming (SQP) on the benchmark, as well as new stochastic and incremental optimization methods borrowed from the large-scale machine learning literature. Interestingly we show that some of these stochastic and incremental methods, which are based on stochastic gradient descent (SGD), achieve higher success rates than SQP on tough initializations.
Inspired by this observation we also propose a new incremental variant of SQP which updates only a random subset of the costs and constraints at each iteration. The algorithm is the best performing in both success rate and convergence speed, improving over SQP by up to 30\% in both criteria.

The benchmark's resources and a solution evaluation script are made openly available.

\end{abstract}

\section{INTRODUCTION}

Optimization of robot postures and trajectories is an important problem in robotics, with applications to efficient robot locomotion and manipulation.
The problem is hard because posture and trajectory optimization problems are often non-convex and have several local optima, especially in the presence of collision constraints. For example when a robot model is in collision, the collision gradient might pull different links in different directions - and the optimization get stuck at an infeasible local optima \cite{Schulman2014}.
While several algorithms have been proposed for posture generation and trajectory optimization \cite{Kuindersma2016,Posa2014,Schulman2014,Zucker2013,Kalakrishnan2011,Lengagne2013,Escande2016}, comparing them is also difficult since each researcher opts for a different evaluation scenario which was chosen by chance or to showcase the advantages of the algorithm.

This paper is a step towards solving these two issues: 1) of the lack of a benchmark, and 2) of avoiding infeasible local optima.
The contributions of the paper are as follows.
\begin{itemize}
	\item We develop a new robotics challenge, open to the public, which consists of a set of problems for benchmarking static posture and trajectory optimization algorithms on a legged humanoid robot
	\item We evaluate different optimization algorithms at solving the problems
	\item We show that new stochastic optimization algorithms developed for the (highly non-convex) training of deep neural networks also lead to high success rates in posture/trajectory optimization with tough collision constraints and initializations. To the best of our knowledge this is the first time these methods are applied to posture generation and trajectory optimization.
	\item We propose a stochastic incremental variant of Sequential Quadratic Programming for trajectory optimization based on the principles of recent large-scale non-convex optimization methods. We show it leads to higher success rates and speed than SQP on ill-initialized trajectory optimization problems.
\end{itemize}

\section{RELATED WORK}

Robot motion planning has been tackled with search, sampling and optimization methods.
Recently, optimization algorithms have gained popularity, due to the existence of fast general-purpose software and the possibility to easily integrate many different constraints in the problem.
One of the state-of-the-art algorithms is sequential quadratic programming, which is used by SNOPT \cite{Gill2002} for general-purpose optimization, but also by trajectory-optimization libraries \cite{Schulman2014}, and trajectory optimization research on legged robots \cite{Posa2014}.
Projected conjugate-gradient methods such as CHOMP \cite{Zucker2013} have also been proposed for the problem. The gradient-descent methods we evaluate in this paper are related to CHOMP in the sense that they also do (pre-conditioned) gradient descent. However, as in \cite{Schulman2014} we use penalties for constraints, thus being able to consider general non-linear constraints on robot postures or motion.

In this paper we explore the use of stochastic methods for posture generation and trajectory optimization. The motivation behind it is to improve success rates by successfully avoiding local minima through random perturbations. The idea has also been explored in stochastic variants of CHOMP \cite{Kalakrishnan2011} which increased success rates. 
Here we instead make use of progress in the stochastic optimization literature, which has recently gained attention in part because of the problem of local minima, saddle points and non-convexities which pervade deep neural network training landscapes.
The large-scale optimization and deep learning communities have recently come up with different algorithms to deal with these optimization landscapes, such as variants of stochastic gradient descent with pre-conditioning \cite{Kingma2014adam}, incremental gradient descent methods \cite{Schmidt2017}, noise injection \cite{Neelakantan2015} and others. Some of these algorithms have provable convergence guarantees \cite{Schmidt2017} and saddle-escaping guarantees \cite{Anandkumar2016}.
Our assumption in this paper is that these methods and insights which work on the highly-nonconvex optimization landscapes of neural networks will transfer to the (also non-convex) landscapes of legged robot posture generation and trajectory optimization.

Results of state-of-the-art robot motion planning algorithms are impressive \cite{Posa2014,Kuindersma2016,Brandao2016planning}, but it is arguably difficult to compare each planner's performance, advantages and disadvantages. This is partly because each algorithm is evaluated on a different environment, or using different cost functions, constraints or robot models.
For results to be comparable and verifiable, the evaluation criteria must be the same and largely sampled, while all inputs (i.e. scenario, robot, constraints) must be available. 
Recently, verifiability and comparability have been strongly pursued in fields such as computer vision through an investment in open benchmarks \cite{Russakovsky2015,Everingham2010} and open source - which has arguably been a strong factor in fostering research progress.
This paper tries to follow this trend and make public a benchmark with pre-defined robot, environment, costs, constraints and evaluation criteria.

\section{THE LegOpt BENCHMARK}

\subsection{The environment}

The LegOpt Benchmark (Legged-robot posture and trajectory Optimization Benchmark), which we propose in this paper, is based on the ``Destruction Scenario Dataset'' (DSD)%
\footnote{The first author of this paper collaborated with $\mu$Roboptics, Inc. on the development of the Destruction Scenarios Dataset.}
\cite{DestructionScenarios}.
DSD is a set of challenging 3D scenarios intended for testing robot search and rescue missions. They were created in Blender using simple shape primitives, unions and differences, followed by shattering and collapsing by running a physics simulation. The scenarios also include 3D models of cars and people for realism. Snapshots of the first frames were taken as different ``difficulty levels'' of the scenarios according to an empirical measure of difficulty of traversing the terrain. Models and code to generate them are openly available%
\footnote{\scriptsize\url{https://github.com/roboptics/destruction_scenarios}}.

\begin{figure}
	\centering
	\includegraphics[width=0.98\linewidth]{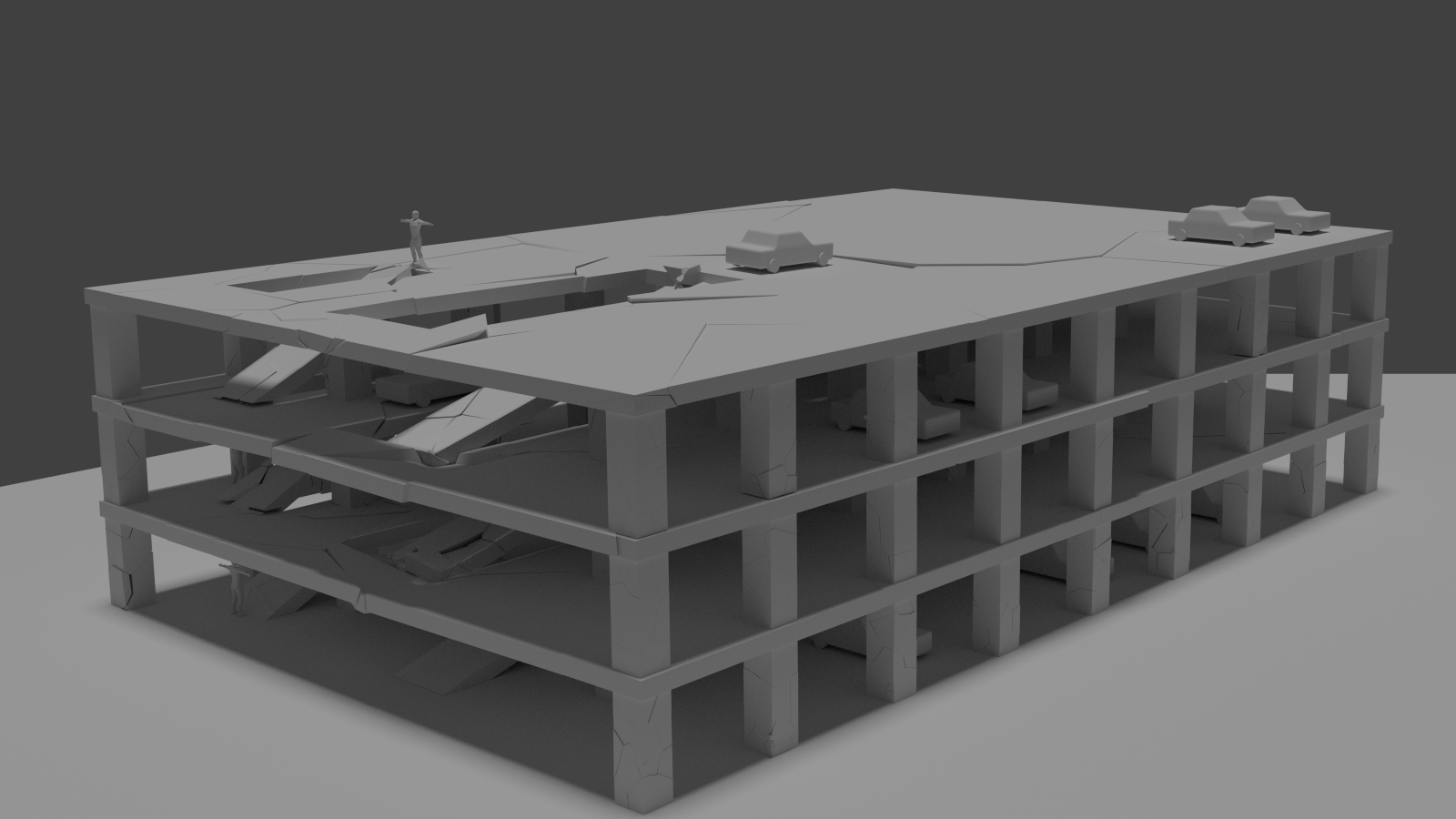}
	\caption{The ``garage\_easier'' scenario: a damaged garage scenario of the Destruction Scenarios Dataset, used as the environment of our benchmark.}
	\label{fig:environment}
\end{figure}

The current version of the LegOpt Benchmark is built on top of the easiest difficulty garage scenario (i.e. ``garage\_easier''). We show an image of the scenario in Figure \ref{fig:environment}. 
This scenario is already challenging enough for current motion planning and optimization algorithms.

\subsection{Task definitions}

The LegOpt Benchmark consists of two tasks: posture generation and trajectory optimization. 
Each task is evaluated in 50 different problem instances (i.e. different regions of the scenario).
Each posture generation problem is specified by a \emph{single stance} with given 6D poses for the robot's feet.
Each trajectory optimization problem is specified by a \emph{sequence of stances} with given feet poses.

Both tasks use Atlas as the robot model and DSD's ``garage\_easier'' scenario as the environment.
The following constraints need to be satisfied for both tasks:
\begin{itemize}
	\item Given feet positions and orientations must be respected,
	\item The center-of-mass (COM) of the robot should lie inside the support polygon, given by a 2D convex hull of the horizontal projection of the feet in contact,
	\item The robot model should not collide with itself or the environment,
	\item Robot joint angle limits should be respected.
\end{itemize}
Each of these constraints is to be enforced with tolerance $10^{-3}$ (meters for distances, radians for angles). We compute collision as the penetration distance in meters, as implemented in the OpenRAVE library \cite{Diankov2010} using ODE \cite{ODE} for collision computations\footnote{We ignore collisions between the pelvis and upper-torso, since their convex hulls are always in collision and most collision checkers make use of convex hulls.}. We provide an evaluation script which computes the costs and constraint violations of JSON-formatted solutions on a website\footnote{\scriptsize\url{https://github.com/martimbrandao/legopt-benchmark}\footnotesize}. We also provide JSON files describing the feet poses of each problem instance.

We evaluate an algorithm's performance at the tasks using the following criteria: 1) Success rate, i.e. number of problem instances solved to tolerance; 2) Cost; and 3) Computation time. The cost function is different for the two tasks.

\subsubsection{Posture generation}
\emph{The posture generation task is to obtain a single full-body configuration $q^*$ that respects given feet poses, static stability, no-collision, joint limits, and leads to the minimum (squared) static torques possible.}
Basically, the task is to solve
\begin{align}
q^* = &\underset{q}{\text{argmin}} && \textstyle\sum\limits_{k=1}^{K} \tau_k^2(q) \\
&\text{s.t.}	&& \text{foot}_i(q) = b_i \,\,\forall i \\
&				&& \text{COM}(q) \in \mathcal{P}_{\text{support}} \\
&				&& \text{sd}(q) \geq 0 \\
&				&& q_{\text{min}} \leq q \leq q_{\text{max}} .
\end{align}
where $K$ is the number of the robot's joints, $b_i$ are the problem's given feet poses, $\mathcal{P}_{\text{support}}$ is the convex support polygon, $\text{sd}(q)$ represents a signed-distance function between the robot, itself and the environment.

\subsubsection{Trajectory optimization}
\emph{The trajectory optimization task is to obtain a sequence of full-body configurations $x^*=q_1^*,...,q_T^*$ (two configurations per stance) which respects given feet poses, static stability, no-collision, joint limits, and leads to the minimum (squared) joint velocity possible.}
Basically, the task is to solve
\begin{align}
q_1^*,...,q_T^* = &\underset{q_1,...,q_T}{\text{argmin}} && \textstyle\sum\limits_{t=1}^{T-1} ||q_{t+1}-q_{t}||^2 \\
&\text{s.t.}	&& \text{foot}_{i,t}(q_t) = b_{i,t} \,\,\forall i,t \\
&				&& \text{COM}_t(q) \in \mathcal{P}_{\text{support}_t} \,\,\forall t \\
&				&& \text{sd}(q_t) \geq 0 \,\,\forall t \\
&				&& q_{\text{min}} \leq q_t \leq q_{\text{max}} \,\,\forall t .
\end{align}

\subsection{Generation of problem instances}

For completeness we will now describe the procedure we used to arrive at the definition of all problem instances. We first sampled points uniformly from the scenario's mesh triangles to obtain positions for a left foot, then we uniformly sampled yaw orientations, which together with the mesh's normal vector make an orientation constraint. We then searched for a right foot position and orientation in the same way, in a radius around the left foot. We checked for feasibility and collision of the stance using TrajOpt. This was done until we had 100 stances scattered throughout the scenario. We then selected random pairs of stances (50 pairs on the same floor, 50 on different floors), and ran an ARA*-based \cite{Likhachev2003} footstep planner \cite{Brandao2016tro} to obtain a sequence of stances which connects the pair. The sequence of stances from the first 25 successful same-floor footstep plans, and the first 25 successful different-floor footstep plans were selected as the problem instances for trajectory optimization. Regarding posture generation, we used K-means clustering to pick 50 stances out of the set of all stances in all footstep plans (using stance length, stance height and trunk collision distance as features).

\begin{figure*}
	\centering
	\includegraphics[width=0.195\linewidth]{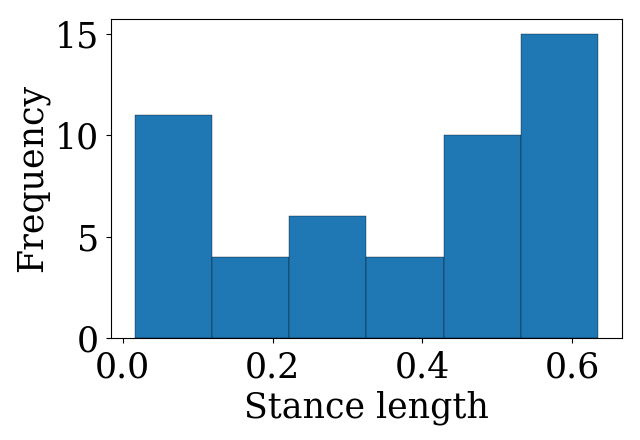}
	\includegraphics[width=0.195\linewidth]{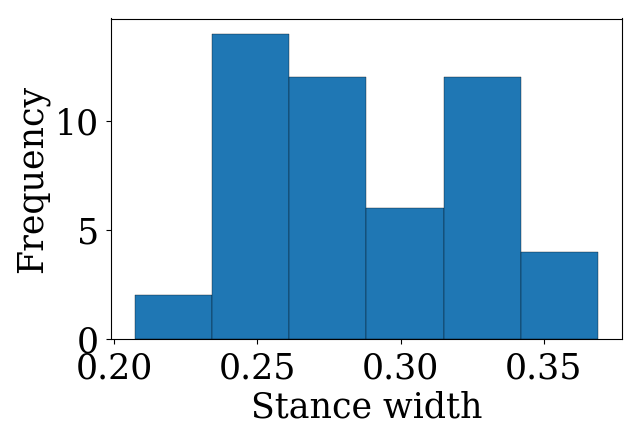}
	\includegraphics[width=0.195\linewidth]{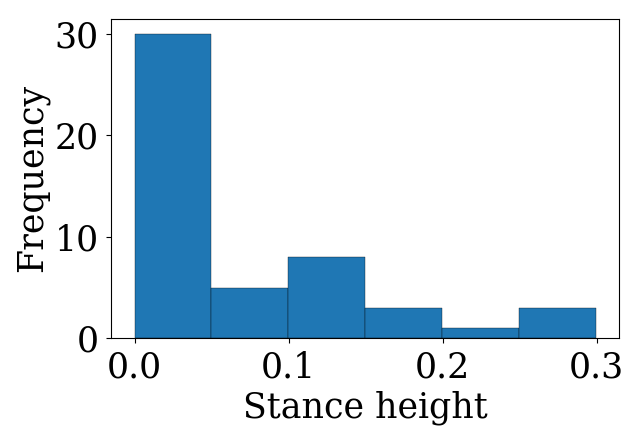}
	\includegraphics[width=0.195\linewidth]{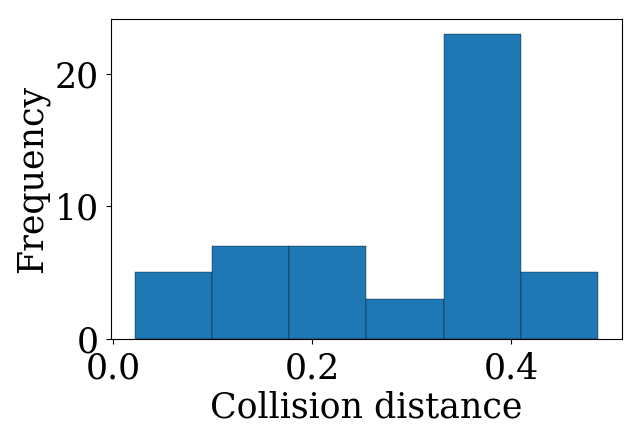}
	\caption{Statistics of the posture generation task: histograms of stance definitions and collision distances on all 50 problems.}
	\label{fig:stats-posture}
\end{figure*}

\begin{figure*}
	\centering
	\includegraphics[width=0.195\linewidth]{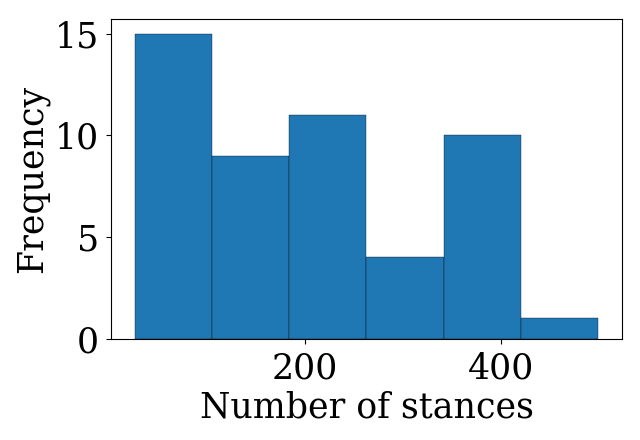}
	\includegraphics[width=0.195\linewidth]{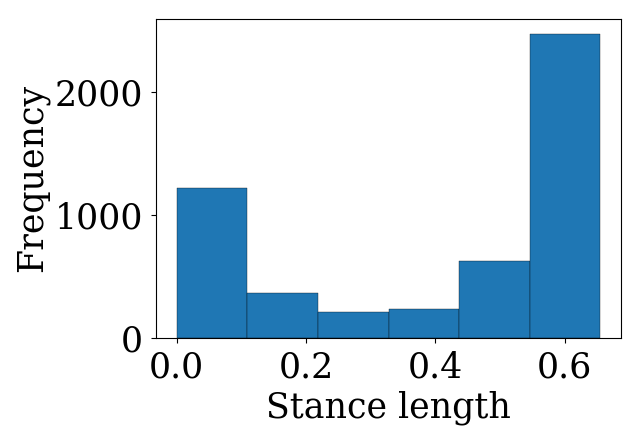}
	\includegraphics[width=0.195\linewidth]{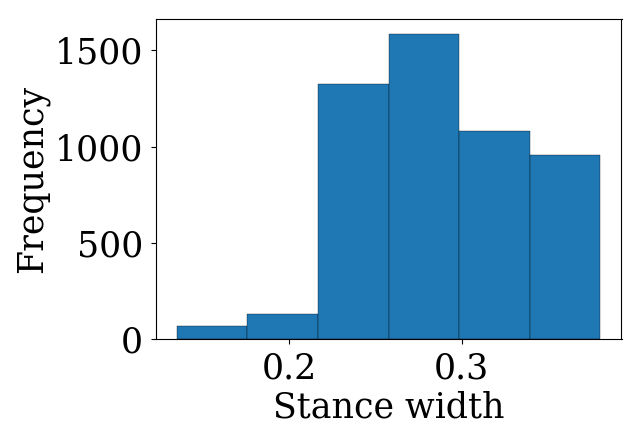}
	\includegraphics[width=0.195\linewidth]{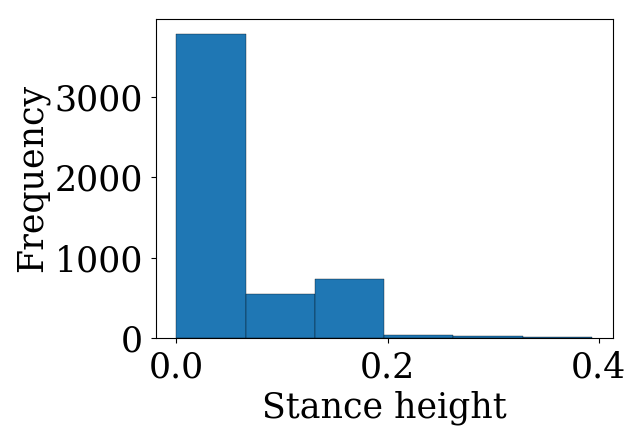}		\includegraphics[width=0.195\linewidth]{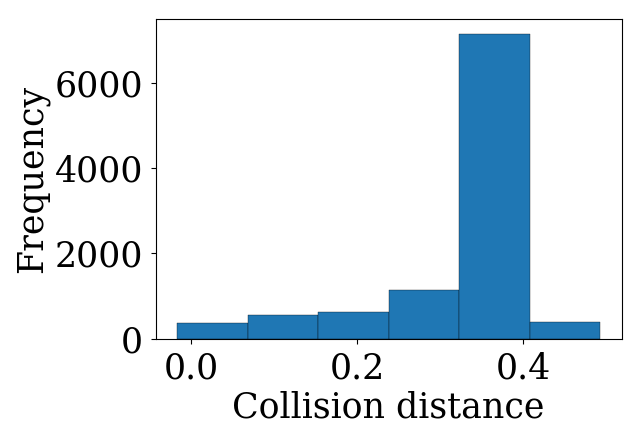}
	\caption{Statistics of the trajectory optimization task: histograms of stance definitions and collision distances on all 50 problems.}
	\label{fig:stats-motion}
\end{figure*}

We show the distribution of problem definition features in Figure \ref{fig:stats-posture} and \ref{fig:stats-motion}. The collision distance feature is computed by a signed-distance function between the trunk (approximated by a box) and the environment. Notice that many postures in both tasks are close to collision, and some will actually be in collision depending on the initialization - since only the trunk and not the arms were checked for collision while building the problem definitions.

\section{BENCHMARKED ALGORITHMS}

Posture generation and motion planning can be formulated as optimization problems of the form
\begin{subequations}
	\begin{align}
	&\underset{x\in\mathbb{R}^d}{\text{min}} 
					&& \textstyle\sum\limits_{i=1}^{n_f} f_i(x) \\
	&\text{s.t.}	&& g_i(x) \leq 0, i = 1,2,...,n_g \\
	&				&& h_i(x) = 0, i = 1,2,...,n_h \\
	&&& \text{linear constraints} ,
	\end{align}
	\label{eq:optoriginal}
\end{subequations}
where $x$ are the full-body postures of the robot at one or more instants of time (i.e. $d=D \times T$ where $D$ is the number of the degrees of freedom of the robot and $T$ the number of instants of time a.k.a. waypoints). Each $f_i$, $g_i$, $h_i$ is a function which describes part of the costs or constraints on the robot, for example a full-body posture cost or a link pose constraint at a certain waypoint.
In this paper we turn (\ref{eq:optoriginal}) into a sequence of problems of the form
\begin{subequations}
	\begin{align}
	&\underset{x\in\mathbb{R}^d}{\text{min}} && \textstyle\sum\limits_{i=1}^{n_f} f_i(x) 
		+ \mu \textstyle\sum\limits_{i=1}^{n_g} p_g(g_i(x)) 
		+ \mu \textstyle\sum\limits_{i=1}^{n_h} p_h(h_i(x)) 
	\label{eq:optpenalty_objective} \\
	&\text{s.t.} && x \in C \quad \text{(convex linear constraints)} ,
	\label{eq:optpenalty_constraint}
	\end{align}
	\label{eq:optpenalty}
\end{subequations}
which are solved with successively higher $\mu$ until all constraints are satisfied (or progress is no longer possible). The functions $p_g(.)$ and $p_h(.)$ are called penalty functions and measure constraint violation. We use $p_g(.)=\text{max}(0,.)$ and $p_h(.)=|.|$, which are exact - meaning that they can lead to the exact solution to the original problem (\ref{eq:optoriginal}) for certain choices of $\mu$ \cite{Nocedal2006}.

\subsection{Sequential quadratic programming (SQP)}

One of the state-of-the-art methods for solving (\ref{eq:optoriginal}) is TrajOpt \cite{Schulman2014}. The method uses sequential quadratic programming, by alternating between 
\begin{enumerate}
	\item building convexifications $\tilde{f}_i$, $\tilde{g}_i$, $\tilde{h}_i$, where $\tilde{f}_i$ are quadratic and $\tilde{g}_i$, $\tilde{h}_i$ are linear
	\item solving (\ref{eq:optpenalty}) with the convexified $\tilde{f}_i$, $\tilde{g}_i$, $\tilde{h}_i$ instead of the original $f_i$, $g_i$, $h_i$ functions using a trust-region method (i.e. constraining $x$ at each iteration to be within a narrow range where convexification is valid). TrajOpt also uses $p_g(.)=\text{max}(0,.)$ and $p_h(.)=|.|$
\end{enumerate}
An implementation of this algorithm is openly available\footnote{\scriptsize\url{http://rll.berkeley.edu/trajopt/}} and achieves both a high success rate and computation speed when compared to other recent optimization-based planners \cite{Zucker2013} and sampling-based planners. 

\subsection{Gradient descent}

A simpler way to solve (\ref{eq:optpenalty}) is to use gradient descent. 
For convenience, let us simplify the notation of the optimization problem (\ref{eq:optpenalty}) as
\begin{equation}
	\underset{x \in C}{\text{min}} \quad 
	\textstyle\sum\limits_{i=1}^{n} F_i(x) ,
	\label{eq:optpenalty2}
\end{equation}
where $n=n_f+n_g+n_h$, and where the terms $F_i(x)$ gather all cost functions and penalties in (\ref{eq:optpenalty_objective}):
\begin{equation}
	F_i(x) =
	\begin{cases}
	f_i(x),					& 1 \leq i \leq n_f \\
	p_g(g_{i-n_f}(x)),		& n_f < i \leq n_f+n_g \\
	p_h(h_{i-n_f-n_g}(x)),	& n_f+n_g < i \leq n .
	\end{cases}
\end{equation}

Gradient descent solves (\ref{eq:optpenalty2}) by iterating
\begin{equation}
	x_{k+1} = 
		\pi_C \left( x_{k} - \alpha_{k} 
		\left( \sum\limits_{i=1}^{n} \nabla F_i(x_{k}) \right)
		\right) ,
	\label{eq:gd}
\end{equation}
where $\pi_C$ projects points onto the feasible space $C$, and $\alpha_{k}$ is called the step length. The step length is usually fixed or computed through line search \cite{Nocedal2006}. The iterations are repeated until some termination criteria is met, usually a threshold on gradient norm or function improvement.
Notice that the gradient of $\text{max}(0,y)$ and $|y|$ is not defined at $y=0$, and so computing (\ref{eq:gd}) with these penalty functions is not straightforward. In this paper we use smooth approximators\footnote{Another option is to use slack variables \cite{Nocedal2006} as done in TrajOpt \cite{Schulman2014}. We chose to use smooth approximators as it made debugging more intuitive.} for the penalty functions such that:
\begin{subequations}
	\begin{align}
	\nabla \tilde{p}_h(h_i(x)) = &\left( -1 + 2 \left( \frac{e^{wh_i(x)}}{1+e^{wh_i(x)}} \right) \right) \nabla h_i(x) \label{eq:smoothpenaltyh} \\
	\nabla \tilde{p}_g(g_i(x)) = &
	\begin{cases}
		0, 							&g_i(x) \leq 0 \\
		\nabla \tilde{p}_h(g_i(x)),	&\text{otherwise}.
	\end{cases} \label{eq:smoothpenaltyg}
	\end{align}
	\label{eq:smoothpenalties}
\end{subequations}
Note that as $w\to\infty$, (\ref{eq:smoothpenaltyh}) tends to $-\nabla h_i(x)$ for $h_i(x)<0$ and tends to $\nabla h_i(x)$ for $h_i(x)>0$, thus approximating the gradient of the exact penalty $|h_i(x)|$; and similarly (\ref{eq:smoothpenaltyg}) also approximates the gradient of $\text{max}(0,g_i(x))$.

\subsection{Popular algorithms in large-scale optimization and deep learning}

\subsubsection{Stochastic gradient descent}

Stochastic gradient descent (SGD) is gradient descent where the gradients are noisy. In this situation an estimator $\tilde{\nabla} F(x_{k})$ is used that is equal to the real gradient in expectation, i.e. $E[\tilde{\nabla} F(x_{k})] = \nabla F(x_{k})$. A common choice for this estimator is the use of mini-batches, which basically evaluates the gradient at only a subset of the functions at each iteration. So the sum over $F_i$ is made not for $i=1,...,n$, but only for a subset $\mathcal{F}_k$ of those indices chosen randomly at iteration $k$. An SGD iteration to solve (\ref{eq:optpenalty2}) then consists of
\begin{equation}
	x_{k+1} = 
	\pi_C \left( x_{k} - \alpha_{k} 
	\left( \sum\limits_{i \in \mathcal{F}_k} \nabla F_i(x_{k}) \right)
	\right) .
\label{eq:sgd}
\end{equation}
The size of a mini-batch ($|\mathcal{F}_k| \leq n$) is usually called ``sample-size'', and for $|\mathcal{F}_k|=n$ SGD reduces to full deterministic gradient descent (\ref{eq:gd}). 
SGD has low complexity (i.e. low total number of gradient evaluations) and provable convergence in probability even on non-convex functions \cite{Bottou2016}, while its noise actually helps avoid sharp local minima \cite{Keskar2016}. Due to these advantages the method has inspired the development of a number of different variants, which try to deal with important disadvantages such as slow convergence and hard-to-tune step lengths. 

\subsubsection{Adam}

In this paper we will consider one of those variants which has become widely popular in the deep learning community: Adam \cite{Kingma2014adam}. Adam basically does SGD where the step length is adapted to each dimension of $x$, similarly in purpose to conjugate gradient descent. In a nutshell, Adam iterates:
\begin{equation}
x_{k+1} = \pi_C \left( x_{k} - \alpha_{k} \frac{\hat{m}_k}{\sqrt{\hat{v}_k}+\epsilon} \right) ,
\label{eq:adam}
\end{equation}
where the division is element-wise, and $\hat{m}_k$, $\hat{v}_k$ are un-biased running averages of the first and second moments of $\tilde{\nabla} F(x_{k})$, accumulated at each iteration the following way
\begin{subequations}
	\begin{align}
		\hat{m}_k = & m_k / (1-\beta_1^k) \label{eq:adam_m} \\
		\hat{v}_k = & v_t / (1-\beta_2^k) \label{eq:adam_v} \\
		m_k       = & \beta_1 m_{k-1} + (1-\beta_1) g_k \\
		v_k       = & \beta_2 v_{k-1} + (1-\beta_2) g_k^2 \\
		g_k       = & \sum\limits_{i \in \mathcal{F}_k} \nabla F_i(x_{k}) .
	\end{align}
\end{subequations}
The constants $\beta_1$ and $\beta_2$ are parameters. We refer the interested reader to the original publication for details. 

\subsubsection{Nadam}

We also consider an accelerated version of Adam, Nadam \cite{Dozat2016}, which uses Nesterov momentum in an attempt to increase convergence rate. The algorithm uses a slightly different definition for $\hat{m}_k$,
\begin{equation}
	\hat{m}_k^{\text{(Nadam)}} = \beta_1 m_k / (1-\beta_1^{k+1}) + (1-\beta_1) g_k / (1-\beta_1^k) .
	\label{eq:nadam_m}
\end{equation}

\subsubsection{Incremental SGD (a.k.a. stochastic average gradient)}

Stochastic average gradient, proposed in \cite{Schmidt2017}, is an incremental variant of SGD. The gradients of all functions are used at each iteration, but only a subset of them is actually computed: the rest are kept unchanged from the previous iteration. The method has faster convergence properties in convex functions than SGD, at the cost of increased memory usage.
It iterates
\begin{equation}
	x_{k+1} = 
	\pi_C \left( x_{k} - \alpha_{k} 
	\left( \sum\limits_{i=1}^n y_i^k \right)
	\right) ,
	\label{eq:sag}
\end{equation}
where
\begin{equation}
	y_i^k =
	\begin{cases}
	\nabla F_i(x_{k}),	& i \in \mathcal{F}_k \\
	y_i^{k-1},			& \text{otherwise}.
	\end{cases}
\end{equation}
As before, the set $\mathcal{F}_k$ is chosen randomly at each iteration.

In the experimental section we will also consider incremental versions of Adam and Nadam (i.e. by keeping previous gradients in memory). For a shared and readable naming, we will refer to all these variants by the prefix "I-", for incremental: I-SGD, I-Adam, I-Nadam.%
\footnote{Note that I-SGD and stochastic average gradient refer to the same method in this paper.}

\subsection{A new, incremental SQP (I-SQP)}

In the spirit of I-SGD, we propose to use an incremental version of SQP. The motivation is to observe that a large part of the computational time of SQP goes into building convexifications (i.e. basically taking the gradients of $f_i$, $g_i$, $h_i$), and so convergence rate can hypothetically be increased by re-using them from previous iterations - as long as the convexifications remain valid within tolerance. 
Another possible motivation is that noise arising from re-using previous convexifications may help avoid infeasible stationary points, as happens with SGD. 

We show the pseudo-code for I-SQP in the following Algorithm \ref{alg:isqp}. It is basically TrajOpt \cite{Schulman2014} with incremental convexifications of costs and penalties.

\begin{algorithm}
	\caption{I-SQP}
	\label{alg:isqp}
	\begin{algorithmic}
		\STATE \textbf{input:} $x$; tolerance, penalty and trust region parameters
		\FOR{PenaltyIteration $\mu \gets \mu_1,\mu_2,...$}
			\FOR{ConvexifyIteration $k \gets 1,2,...$}
				\IF{$k=1$}
					\STATE $\tilde{F}_i^k \gets$ Convexify($F_i$) $\forall i=1,...,n$
				\ELSE
					\STATE $\mathcal{F}_k \gets$ SampleMiniBatch()
					\STATE $\tilde{F}_i^k \gets$ Convexify($F_i$), $\forall i \in \mathcal{F}_k$
					\STATE $\tilde{F}_i^k \gets \tilde{F}_i^{k-1}$, $\forall i \notin \mathcal{F}_k$
				\ENDIF
				\STATE Using a trust-region algorithm, solve:
				\STATE $\quad x \gets \underset{x \in C}{\text{argmin}} \sum\limits_{i=1}^{n} \tilde{F}_i^k(x)$
				\IF{constraints satisfied to tolerance}
					\STATE break
				\ENDIF
			\ENDFOR
		\ENDFOR
	\end{algorithmic}
\end{algorithm}

\section{RESULTS}

\subsection{Experimental setup and implementation}

Performance of each optimization method varies greatly with the choice of initialization, especially in the presence of non-convex constraints such as obstacle avoidance. Optimizer behavior close and far from collision may also vary greatly, and therefore be informative of the advantages and disadvantages of each method.
Because of this, in this paper we obtain results for two different initializations:
\begin{enumerate}
	\item ``Good'' initialization: the robot's base is placed such that its feet are above their target pose (hence not in collision with the ground). In particular, the optimization will start from $q$ such that the feet soles' height will be equal to the maximum target foot sole height plus a margin of 0.20 meters. The joints are always set to a nominal posture (knees bent 50 degrees), and the base's XY position is set to the center of the active contacts.
	\item Initialization ``in-collision'': the robot's base is placed such that its feet are in collision with the ground. In particular, the optimization will start from $q$ such that the feet soles' height will be equal to the minimum target foot sole height minus a margin of 0.05 meters. The joints are always set to a nominal posture (knees bent 50 degrees), and the base's XY position is set to the center of the active contacts.
\end{enumerate}

As is common in optimization methods for robotics, we use ``multi-starts'' (a.k.a. restarts) to deal with optimization failure. This means that in case the optimization does not converge from a certain initialization, we restart it from scratch using a slightly perturbed initial condition. In our experiments we restart the optimization up to 10 times, and each time perturb all joint angles using a uniform distribution in the interval of $[-5,5]$ degrees.
We also report results without using multi-starts, although they have expectedly lower success rates.

We solved all problems with full, stochastic and incremental gradients, with sample sizes $20,40,...,100\%$ of $n$. On average we obtained best-performing results using $80\%n$ for stochastic methods, $40\%n$ for incremental gradient methods and $80\%n$ for I-SQP. For brevity we report results using these sample sizes only.
Method SQP with $80\%n$ is equivalent to I-SQP but without memory (i.e. not incremental). Although we do not expect the algorithm to score high, because each QP will be very different from the next as certain constraints are removed, we still include it in the comparison for thoroughness.

We used cost and constraint implementations directly from the TrajOpt library \cite{Schulman2014}. Although it is not part of the trajectory optimization task's requirements, we also included a continuous collision constraint between waypoints to obtain trajectories that are closer to interpolation-feasible.

Due to the large number of variables on the trajectory optimization task, we solve each trajectory optimization problem as a sequence of optimization problems which consider only 6 stances (12 waypoints), with an overlap of 1 stance that is re-planned. The first of those waypoints is fixed (planned in the previous window).

We implemented all optimization methods in C++ as extensions to the TrajOpt library. The same implementation for the costs, constraints and respective gradient computations is used for all optimization methods - thus making success rates and computation times directly comparable. The only exception is the implementation of penalty function gradients, which in the case of SQP and I-SQP uses slack variables, while other methods use (\ref{eq:smoothpenalties}).

We compute the step-size of each step in gradient-based methods through line-search, using a zoom function and Wolfe's conditions ($c_1=10^{-4}$, $c_2=0.9$), as in \cite{Nocedal2006}. For Adam and Nadam we use standard parameters $\beta_1=0.9$ and $\beta_2=0.999$. The termination criteria is as in \cite{Schulman2014} for SQP methods, and is based on proportional-decrease monitoring for gradient descent methods (i.e. terminate after 10 epochs without improvement in the objective function). Problems are solved successively with penalty coefficients $\mu=10^2,10^3,...,10^{12}$.
We use Bullet for mesh-collision, and solve QPs and feasible-space projections $\pi_C$ using Gurobi.

\subsection{Algorithm comparison}

\begin{figure}
	\centering
	\includegraphics[width=0.70\linewidth]{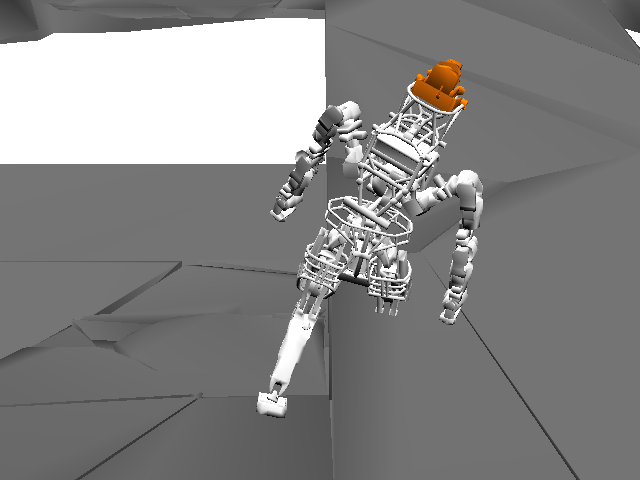}
	\caption{A tough initial condition corresponding to a local minimum.}
	\label{fig:motion-hard-incollision}
\end{figure}

We show all results for both tasks condensed in Table \ref{table:benchmark}.
\begin{table*}
	\begin{center} 
		\caption{ Algorithm comparison on the posture and motion tasks } 
		\label{table:benchmark} 
		\begin{centering}
			\begin{tabular}{ccc|ccc|ccc||ccc|ccc} 
				\multicolumn{3}{c}{}&\multicolumn{6}{c}{Posture generation}&\multicolumn{6}{c}{Trajectory optimization} \\
				\multicolumn{3}{c}{}&\multicolumn{3}{c}{Good-init}&\multicolumn{3}{c}{Init-in-collision}&\multicolumn{3}{c}{Good-init}&\multicolumn{3}{c}{Init-in-collision} \\
				\toprule 
				
				Method     &Multi  &Sample &Success &Cost &Time     &Success &Cost &Time     &Success &Cost &Time     &Success &Cost &Time     \\
				&starts &size   &rate    &     &(s)      &rate    &     &(s)      &rate    &     &(s)      &rate    &     &(s)      \\
				\midrule

				SQP     & 0 &100  &49/50 &427 &1.32 &43/50 &416 &1.43 &19/50 &0.13 &0.63 &11/50 &0.12 &0.67 \\
				SGD     & 0 &100  &44/50 &720 &3.76 &41/50 &770 &3.55 &4 /50 &0.20 &1.38 &2 /50 &0.18 &1.26 \\
				Adam    & 0 &100  &42/50 &606 &4.15 &46/50 &616 &3.78 &18/50 &0.14 &1.22 &2 /50 &0.14 &1.27 \\
				Nadam   & 0 &100  &39/50 &590 &2.83 &48/50 &645 &2.33 &16/50 &0.14 &0.81 &1 /50 &0.11 &0.92 \\
				\midrule
				SQP     &10 &100  &50/50 &427 &1.44 &50/50 &429 &1.74 &42/50 &0.13 &0.62 &29/50 &0.12 &0.84 \\
				SGD     &10 &100  &49/50 &798 &6.30 &48/50 &783 &5.11 &27/50 &0.17 &2.17 &29/50 &0.16 &3.04 \\
				Adam    &10 &100  &49/50 &643 &5.97 &50/50 &672 &4.30 &31/50 &0.14 &1.50 &32/50 &0.15 &2.10 \\
				Nadam   &10 &100  &49/50 &642 &5.67 &50/50 &679 &3.20 &35/50 &0.15 &1.19 &30/50 &0.15 &1.96 \\
				\midrule
				SQP     &10 & 80  &16/50 &298 &2.74 &20/50 &279 &4.00 &0 /50 &0.00 &0.00 &0 /50 &0.00 &0.00 \\
				SGD     &10 & 80  &50/50 &788 &5.82 &49/50 &775 &4.44 &26/50 &0.16 &1.99 &30/50 &0.16 &2.49 \\
				Adam    &10 & 80  &50/50 &629 &4.18 &50/50 &642 &3.91 &35/50 &0.14 &2.05 &36/50 &0.15 &2.12 \\
				Nadam   &10 & 80  &50/50 &620 &2.90 &50/50 &638 &2.70 &33/50 &0.14 &1.10 &34/50 &0.14 &1.47 \\
				\midrule
				I-SQP   &10 & 80  &50/50 &461 &1.00 &50/50 &468 &1.34 &42/50 &0.13 &0.53 &38/50 &0.14 &0.71 \\
				I-SGD   &10 & 40  &49/50 &833 &4.13 &49/50 &810 &5.17 &30/50 &0.15 &1.30 &30/50 &0.15 &1.40 \\
				I-Adam  &10 & 40  &50/50 &661 &3.27 &50/50 &668 &3.66 &40/50 &0.14 &1.07 &34/50 &0.15 &1.34 \\
				I-Nadam &10 & 40  &50/50 &667 &2.66 &50/50 &684 &3.44 &41/50 &0.15 &0.73 &28/50 &0.14 &1.64 \\
				
				\bottomrule 
			\end{tabular}
		\end{centering}
	\end{center}
	\begin{center}
		Note: Costs and times are averaged over successfully solved problems.
	\end{center}
\end{table*}
\begin{figure*}
	\centering
	\includegraphics[width=0.22\linewidth]{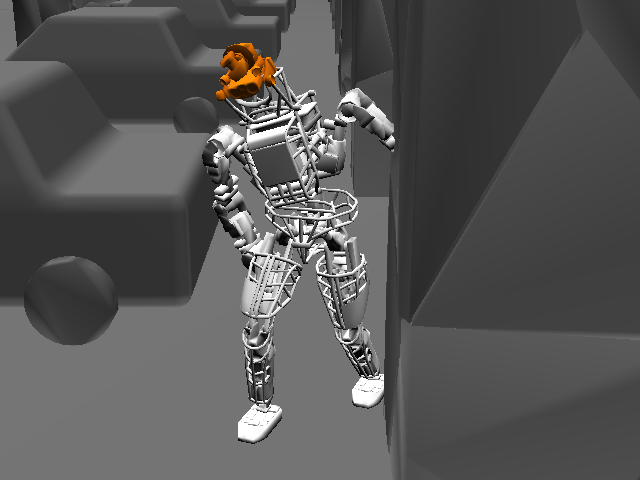}
	\includegraphics[width=0.22\linewidth]{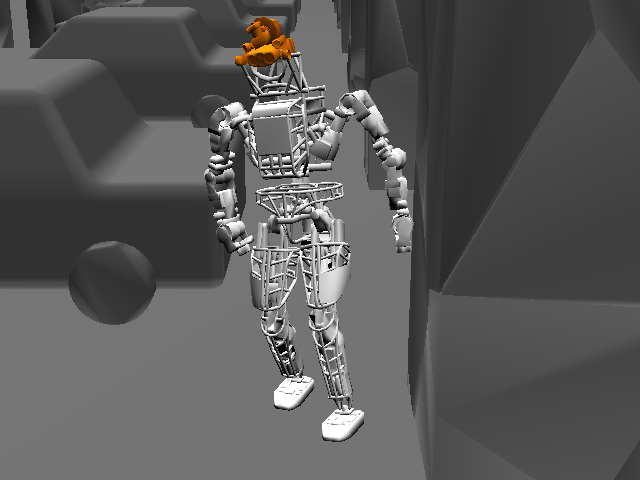}
	\includegraphics[width=0.22\linewidth]{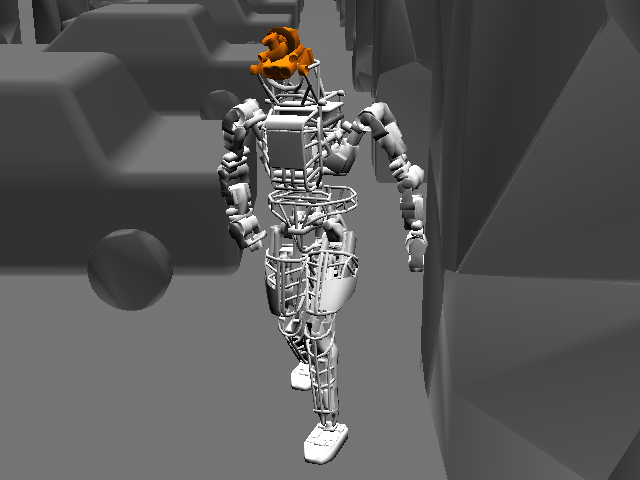}
	\includegraphics[width=0.22\linewidth]{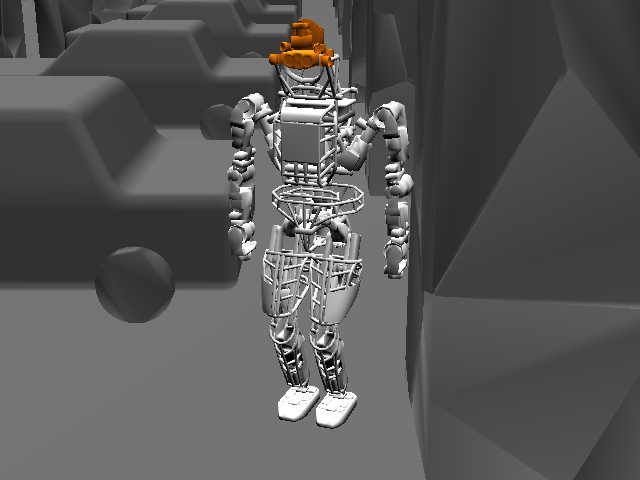}
	\includegraphics[width=0.22\linewidth]{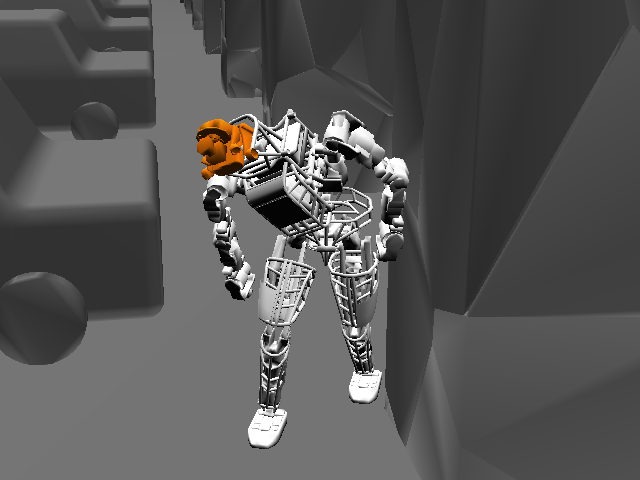}
	\includegraphics[width=0.22\linewidth]{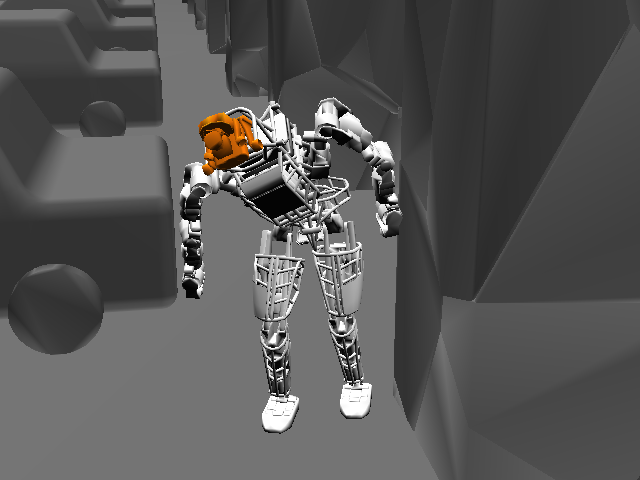}
	\includegraphics[width=0.22\linewidth]{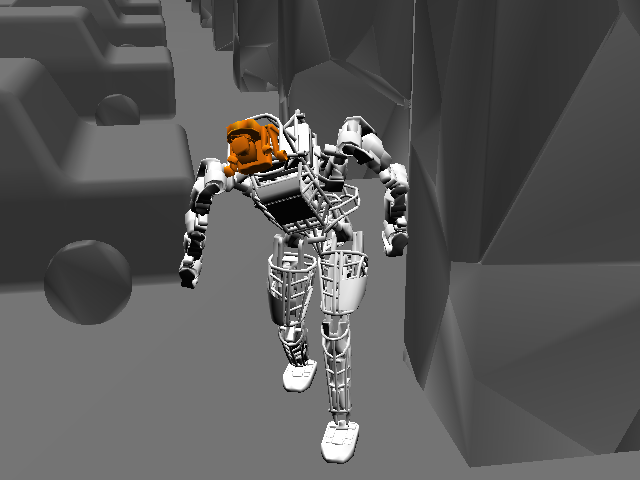}
	\includegraphics[width=0.22\linewidth]{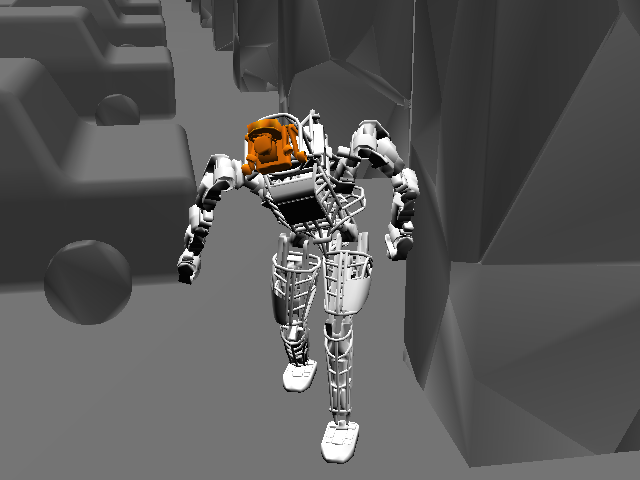}
	\includegraphics[width=0.22\linewidth]{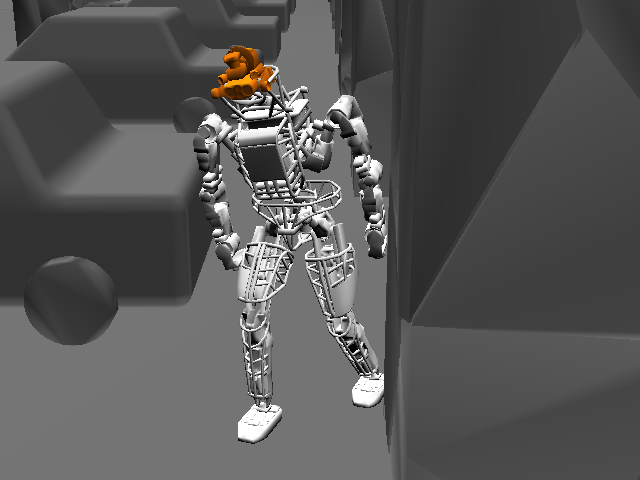}
	\includegraphics[width=0.22\linewidth]{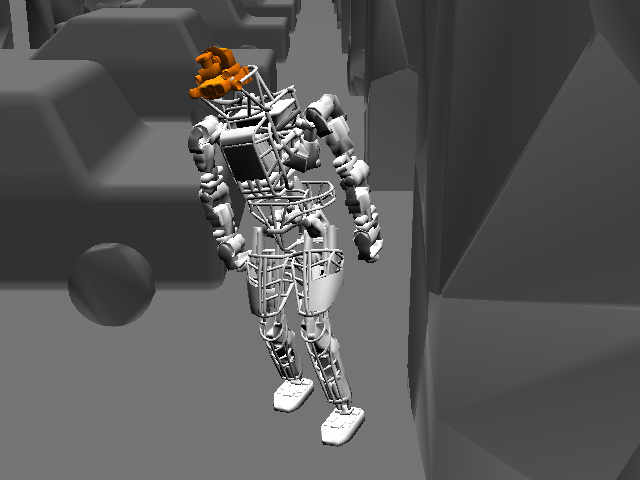}
	\includegraphics[width=0.22\linewidth]{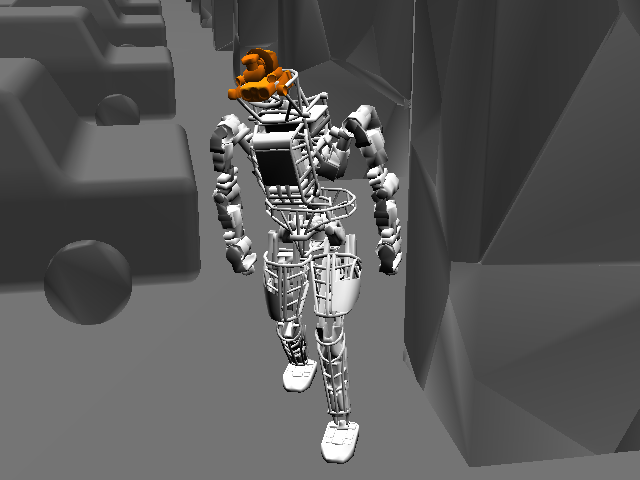}
	\includegraphics[width=0.22\linewidth]{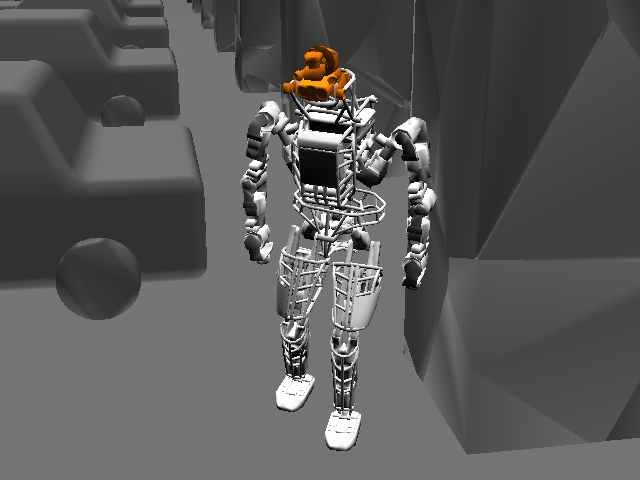}
	\caption{A sequence of waypoints from a trajectory optimization problem. The sequence runs from left to right. Each row corresponds to a different method. From top to bottom: SQP, I-SQP, Adam (80\% batch size).}
	\label{fig:motion31good}
\end{figure*}
As expected, the use of multi-starts greatly improves performance on all methods and conditions.
In general, success rates are higher for the posture generation than the trajectory optimization task. This is understandable because trajectory problems consist of hundreds of postures (see Figure \ref{fig:stats-motion}), while posture generation problems consist of only one.
Success is also higher for the ``good-initialization'' condition compared to the ``in-collision'' condition, which proves the difficulty of optimization methods in dealing with collision constraints. Most of the failures in this condition consist of the optimization method getting stuck at an infeasible stationary point where no progress is made in any degree of freedom (i.e. collision's signed distance gradient is inconsistent between links or cancels out the gradient of other constraints). We show an example of such a case in Figure \ref{fig:motion-hard-incollision}.

Stochastic and incremental gradient-descent methods, as well as our I-SQP, consistently lead to higher success rates than their full-gradient counterparts.
SQP and I-SQP have the lowest computation times, of 1 second per waypoint on average. In fact, I-SQP is 20 to 30\% faster than SQP on average. Gradient-descent variants are slower. I-Adam takes 1 to 3 seconds per waypoint on average, while the Nesterov-accelerated I-Nadam is around 15\% faster on average.
When it succeeds in finding a solution, SQP consistently obtains the lowest-cost postures and trajectories, which we assume to be due to the nonexistence of noise as well as an efficient use of second order information and conditioning employed by the QP solver (Gurobi).
The best performing method in both success rate and computation speed is always I-SQP. Its cost values are similar but slightly higher than SQP.

For the posture generation task, on average all 50 problems are successfully solved by all methods. Computation speed is understandably faster for SQP and I-SQP. This is because they use second-order information and only recompute gradients after convergence of each QP, while gradient methods recompute gradients after each step (i.e. after each line search).
In their stochastic and incremental versions, both SGD- and SQP-based methods decrease computation time on average, some methods obtaining up to 50\% speedups (I-Adam).

\begin{figure}
	\centering
	\includegraphics[width=0.49\linewidth]{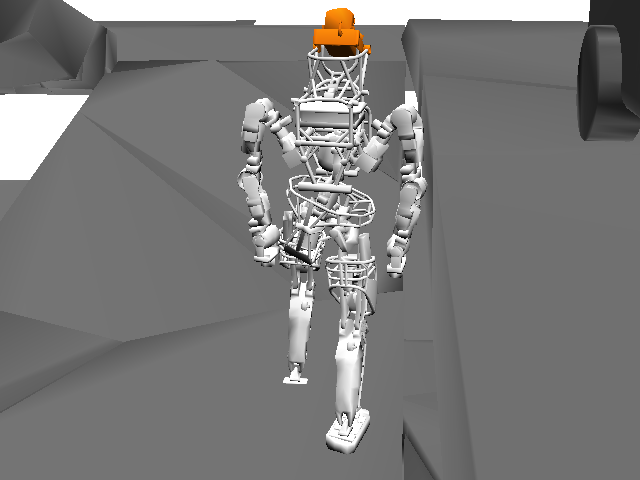}
	\includegraphics[width=0.49\linewidth]{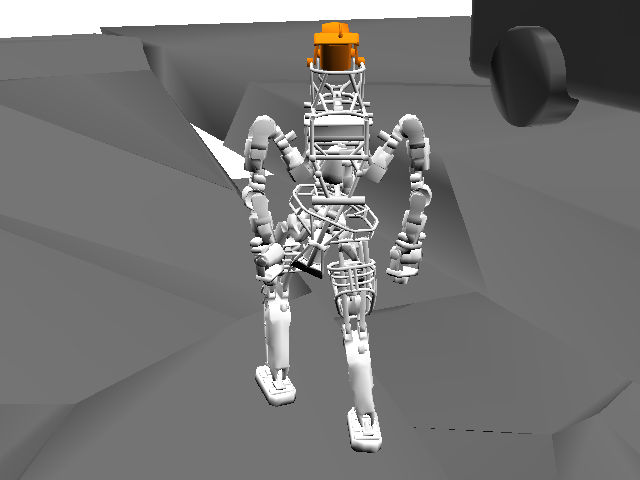}
	\caption{A posture generation problem where SQP fails (left foot is in collision) but I-SQP and incremental/stochastic gradient methods converge. From left to right: SQP, I-SQP.}
	\label{fig:posture-hard}
\end{figure}

The most common reason for algorithm failure is due to collision constraints, an example of which is shown in Figure \ref{fig:posture-hard}. In this situation, SQP fails and requires several random restarts until it finds a feasible solution. I-SQP and all stochastic and incremental gradient methods succeed at first try. The reason is that SQP gets stuck on an infeasible local minimum when a foot is in collision, while noisy gradients in the incremental/stochastic methods help move the robot out of collision (i.e. out of the local minimum).

For the trajectory optimization task, I-SQP is the best performing in both the well-initialized and in-collision conditions, solving 42 and 38 problems respectively. I-SQP considerably improves the success rate with respect to traditional SQP when initialized in-collision (38 vs 29 successes).
I-Adam performs similarly, although slightly worse than SQP on well-initialized problems (40 vs 42). However, I-Adam is better than traditional SQP when started in-collision (34 vs 29). In fact, all stochastic and incremental gradient-based methods outperform traditional SQP in-collision, except for I-Nadam.

Figure \ref{fig:motion31good} shows a sequence of waypoints from a trajectory optimization problem where SQP fails, but I-SQP succeeds, and Adam (sample-size $80\%$) also succeeds. SQP gets stuck on a local minimum where the left arm is in collision with a wall. However, I-SQP successfully finds a posture which brings the arm out of collision, and Adam does so with a simpler posture closer to its nominal value (i.e. straight body with bent knees).



\section{CONCLUSIONS}

In this paper we introduced a new challenge - LegOpt - for benchmarking different algorithms at the tasks of legged robot posture generation and trajectory optimization. LegOpt is open to public and defines a robot model, scenario, constraints and cost functions. Evaluation scripts are also made public for ease of use. LegOpt will hopefully help the legged and humanoid robotics communities to better compare the advantages and disadvantages of planning algorithms on common problems.

We evaluated not only state-of-the-art algorithms based on sequential convex optimization (SQPs), but also recent stochastic and incremental algorithms introduced for large-scale optimization in general and deep learning in particular. The assumption was that their robustness to local minima and saddle points would transfer to the optimization landscapes of robot optimization tasks as well.
The experimental results showed that stochastic and incremental methods are indeed effective in the LegOpt Benchmark, especially on the presence of complex obstacle geometries or on initializations that are in collision. In particular, stochastic and incremental variants of SGD and Adam surpassed SQP when initialized in-collision.
Inspired by these algorithms we proposed an incremental version of SQP for optimizing robot motion, which updates only a random subset of the costs and constraints at each iteration. The algorithm obtained the highest success rates on tough initializations, and was consistently faster than traditional SQP.

Traditional and incremental SQP each performed better in different criteria. SQP arrived at better cost values when it found a solution, while I-SQP was better regarding success rates by avoiding infeasible local minima. 
We believe results can be improved further by the use of continuation methods or more elaborate stochastic methods.

We realize many algorithms have been left out of this benchmark, of notice for example CHOMP and its variants. We hope the community will step forward and report the performance of different algorithms on LegOpt. We will also pursue the evaluation of new methods, with a special interest in stochastic methods.
Although in this paper we define tasks and benchmark algorithms for a biped robot on the easiest-difficulty scenario from DSD \cite{DestructionScenarios}, we envision the extension of the benchmark to multi-contact tasks and tougher DSD scenarios. An extension to locomotion with dynamics is also important future work. The pace of these extensions will depend on progress made in this current version of the benchmark.

\section*{ACKNOWLEDGMENT}

We would like to thank $\mu Roboptics, Inc.$ for developing and opening the Destruction Scenarios Dataset, as well as helping us adapt it to our needs.


\bibliographystyle{IEEEtran}

\end{document}